\documentclass[11pt]{article}
\usepackage[hyperref]{ccl2021-en}
\usepackage{times}
\usepackage{url}
\usepackage{latexsym}
\usepackage{fancyhdr}

\usepackage{amsmath}
\usepackage{xcolor}
\usepackage{multirow}
\usepackage{booktabs}
\usepackage[switch]{lineno}
\usepackage{graphicx}

\title{Enhancing Question Generation with Commonsense Knowledge}

\author{Xin Jia\dag, Hao Wang\ddag, Dawei Yin\ddag, Yunfang Wu\dag\thanks{\quad Corresponding author.} \\
  \dag MOE Key Lab of Computational Linguistics, School of EECS, Peking University \\
  \ddag Baidu Inc., China \\
  {\tt \{jemmryx, wuyf\}@pku.edu.cn}\\
  {\tt way\_wh@yeah.net, yindawei@acm.org}
  }
\date{}

\begin{document}
\maketitle
\begin{abstract}
Question generation (QG) is to generate natural and grammatical questions that can be answered by a specific answer for a given context. Previous sequence-to-sequence models suffer from a problem that asking high-quality questions requires commonsense knowledge as backgrounds, which in most cases can not be learned directly from training data, resulting in unsatisfactory questions deprived of knowledge. In this paper, we propose a multi-task learning framework to introduce commonsense knowledge into question generation process. We first retrieve relevant commonsense knowledge triples from mature databases and select triples with the conversion information from source context to question. Based on these informative knowledge triples, we design two auxiliary tasks to incorporate commonsense knowledge into the main QG model, where one task is Concept Relation Classification and the other is Tail Concept Generation. Experimental results on SQuAD show that our proposed methods are able to noticeably improve the QG performance on both automatic and human evaluation metrics, demonstrating that incorporating external commonsense knowledge with multi-task learning can help the model generate human-like and high-quality questions. 
\end{abstract}

\section{Introduction}
\label{intro}

Question Generation (QG) has become an essential task for NLP, which aims to generate grammatical and fluent questions for a given context and answer. QG can create question-answer pairs as data augmentation for Question Answering (QA)  \cite{Tang2017QuestionAA,Duan2017QuestionGF,Zhang2019AddressingSD}. Moreover, it is also useful in education \cite{Heilman2010GoodQS,Jia2020EQGRACEEQ} and business applications \cite{mostafazadeh-etal-2016-generating}, such as creating materials for language beginners, helping build chatbots, etc.

Existing question generation methods can be roughly grouped into two categories. First, rule-based methods utilize handcrafted paradigms to perform declarative-to-interrogative sentence transformations \cite{Heilman2009QuestionGV,Dhole2020SynQGSA}, but these methods often consume lots of efforts from domain experts and usually cover limited areas. Second, neural network-based methods typically model the question generation task in a fully data-driven manner \cite{Du2017LearningTA,Zhou2017NeuralQG,Wang2020PathQGNQ}, which has made much progress in recent years.

One key issue, however, is still up in the air: human beings often ask questions with commonsense knowledge that may exist in their brain but not appear in the given context. Take the instance in Table \ref{example} as an example. To generate the human-like high-quality question, one must have the corresponding commonsense knowledge that the ``European Parliament" and ``Council of the European Union" are of ``governing bodies". Lacking such commonsense knowledge results in unsatisfactory questions that simply copy some words from the source context. This directs us to introduce commonsense backgrounds to bridge the knowledge gap between the given contexts and generated questions. 

\begin{table}[t]
    \centering
    \scalebox{0.8}{
    \begin{tabular}{p{1.0\columnwidth}}
    \hline
         \textbf{Context Passage:}\\
         The {\color{blue}{European Parliament and the Council of the European Union}} have powers of amendment and veto during the legislative process.\\
         \hline
         \textbf{Reference question:}\\
         Which {\color{red}{governing bodies}} have  legislative veto power?\\
         \textbf{Generated question by the baseline model:}\\
         What has the powers of amendment and veto during the legislative process?\\
         \hline
         \textbf{Extracted commonsense knowledge:}\\
         ({\color{blue}{`council'}}, `RelatedTo', {\color{red}{`governing'}})\\
         ({\color{blue}{`parliament'}}, `Hypernymy', {\color{red}{`legislative bodies'}})\\
    \hline
    \end{tabular}
    }
    \caption{A real example in the training set of SQuAD, which demonstrates the vital effect of commonsense knowledge on QG.}
    \label{example}
\end{table}

Actually, previous NLP works have investigated the structured commonsense knowledge to help text generation, such as story generation \cite{Yang2019EnhancingTG,Guan2019StoryEG} and response generation \cite{Zhou2018CommonsenseKA}. They model commonsense knowledge from external databases as additional context through attention mechanism \cite{Zhou2018CommonsenseKA,Bai2019VariationalAF}. However, simply employing these methods may not perform well in question generation task. Moreover, there remain two open issues in modeling external knowledge for text generation: 1) existing methods \cite{Zhou2018CommonsenseKA,Bai2019VariationalAF} usually utilize all extracted knowledge triples indiscriminately, which neglects the fact that some triples may not provide useful information, and introduce noises. 2) existing methods simply plug the knowledge triples into encoders, which may not fully leverage the information of these knowledge triples.

We here propose more sophisticated modeling on knowledge triples to help question generation: relevant triples, which cover the knowledge gap between source contexts and generated questions, are selected in the very beginning; furthermore, we not only utilize knowledge triples as additional inputs, but also design auxiliary tasks to help the QG model deeply absorb commonsense knowledge. To the best of our knowledge, we are the first to incorporate structured commonsense knowledge into question generation via a multi-task learning framework.

Specifically, we first retrieve all context-relevant knowledge triples from ConceptNet \cite{Speer2017ConceptNet5A} and WordNet \cite{Miller1995WordNetAL}, and keep the triples where the head concept appears in the context and the tail concept appears in the reference question, i.e., (``council", ``RelatedTo", ``governing") and (``parliament", ``Hypernymy", ``legislative bodies") in Table \ref{example}.

Then, we design a multi-task learning framework that combines the main QG task and two triple-based auxiliary tasks: Concept Relation Classification and Tail Concept Generation to benefit the question generation process. The two auxiliary tasks can provide useful knowledge information and optimize the parameters of the main QG model.

We conduct extensive experiments on SQuAD dataset, and our proposed model outperforms strong baselines and achieves comparable state-of-the-art performance, demonstrating that incorporating commonsense knowledge with multi-task learning is able to improve the performance of question generation. We will release our data and code for future research.

\section{Related Work}
Traditionally, QG is tackled by rule-based methods \cite{Heilman2010GoodQS,Labutov2015DeepQW,Dhole2020SynQGSA} that rely heavily on extensive hand-crafted rules. Different from these, neural network-based methods are completely data-driven and trainable in an end-to-end fashion \cite{Du2017LearningTA,Zhou2017NeuralQG,Zhao2018ParagraphlevelNQ,Song2018LeveragingCI,Kim2018ImprovingNQ,Nema2019LetsAA,Zhou2019MultiTaskLW,zhou-etal-2019-question,Jia2020HowTA,Ko2020InquisitiveQG}. For better representing the input context, the answer position and token lexical features (e.g. NER, POS and word case) are treated as supplements for the neural encoder \cite{Zhou2017NeuralQG,Song2018LeveragingCI}. Pointer or copy mechanisms \cite{See2017GetTT,Gu2016IncorporatingCM,Zhao2018ParagraphlevelNQ} are also utilized to overcome the OOV problem in question generation process.

In order to optimize the parameters of QG model, recent works adopt the multi-task learning framework with different auxiliary tasks. Zhou \shortcite{Zhou2019MultiTaskLW} use language modeling as a low-level task to provide coherent representations of the input context for the high-level QG task. To improve the accuracy of the start-up word generation, Zhou \shortcite{zhou-etal-2019-question} treat question type prediction as an auxiliary task and use the predicted word to initialize the decoding process. Jia \shortcite{Jia2020HowTA} acquire built-in paraphrase knowledge through back-translation, and introduce paraphrase knowledge into QG process. Different from these works, this paper introduces external commonsense knowledge into QG through multi-task learning framework. 

In addition to unstructured knowledge as Jia \shortcite{Jia2020HowTA} used, many works in text generation utilize structured knowledge from mature database like ConceptNet \cite{Speer2017ConceptNet5A}. Yang \shortcite{Yang2019EnhancingTG} employ external commonsense knowledge through dynamic memory mechanism to generate more diverse essays. Guan \shortcite{Guan2019StoryEG} apply structured commonsense knowledge through multi-source attention to facilitate story comprehension and generate coherent endings. Zhou \shortcite{Zhou2018CommonsenseKA} incorporate commonsense knowledge graphs through graph attention to create more appropriate and informative responses. Instead of combining knowledge triples into encoding process, our model creatively incorporates commonsense knowledge into QG procedure with two new auxiliary tasks. In this way, the commonsense knowledge can be effectively absorbed by question generation.

\section{Knowledge Extraction}
To incorporate structured commonsense knowledge into question generation, the most basic step is to extract proper knowledge triples for each training sample. We will describe the details of the knowledge extraction in this section.

In order to obtain more commonsense knowledge, we extract structured knowledge triples from two commonly used databases: ConceptNet and WordNet. ConceptNet is a semantic network composed of triples $\mathbf{(h,r,t)}$ denoting that the head concept $\mathbf{h}$ has a relation $\mathbf{r}$ with tail concept $\mathbf{t}$. WordNet is a lexical database organized in accordance with psycholinguistic theories, where lexicalized concepts are organized by semantic relations (synonymy, hyponymy, etc.). For each sample in the training set of SQuAD, we use each non-stop word in the context passage as a query to retrieve corresponding triples from both ConceptNet and WordNet.

In the process of question generation, only those triples that provide essential knowledge for source contexts and target questions are useful rather than all retrieved triples. Therefore, we design a rule to filter triples: keep only those triples where the head concept $\mathbf{h}$ appears in the context passage and the tail concept $\mathbf{t}$ appears in the question (in the case of reversed, we swap the head and tail concepts), since these triples can directly provide conversion information between the input context and the output question. For example, in Table \ref{example}, we can extract triples like (``council", ``RelatedTo", ``governing"), (``council", ``RelatedTo", ``city") and (``council", ``Synonymy", ``assembly") for the word ``council", while we only maintain (``council", ``RelatedTo", ``governing") since it directly provides the needed information for generating the right question.

From ConceptNet, there are 12,432 training samples that have extracted knowledge triples, and from WordNet there are 41,049 training samples successfully extract knowledge triples. For each training sample, we merge these triples from ConceptNet and WordNet and then remove repeated ones. Finally we obtain 46,455 knowledge-equipped samples and each sample has 1.7 corresponding triples on average, as clearly shown in Table \ref{triple_statistics}. 

Accordingly, we divide the origin SQuAD training set into two parts: \textbf{commonsense knowledge-equipped samples}: (context passage, answer, question, knowledge triples) and \textbf{pure samples}: (context passage, answer, question). We will explain how to use these two parts of data in the following sections.

\begin{table}[t]
    \centering
    \scalebox{0.86}{
    \begin{tabular}{c|ccc}
    \hline
         & SQuAD & knowledge-equipped & pure \\
    \hline
    \hline
        ConceptNet & - &12432 &-\\
        WordNet &- &41049 &-\\
        Whole & 75722 & 46455 (61.3\%) &29267 (38.7\%) \\
        
     \hline
    \end{tabular}
    }
    \caption{The statistics of retrieved knowledge triples from ConceptNet and WordNet for the training set of SQuAD. }
    
    \label{triple_statistics}
\end{table}
\begin{table}[t]
    \centering
    \begin{tabular}{cc|cc}
    \hline
         type & proportion & type & proportion  \\
    \hline
    \hline
         Synonymy & 41\% & RelatedTo & 38\% \\
         IsA & 6\%  & Hypernymy & 6\% \\
         Hyponymy  & 3\% & Others & 6\% \\
    \hline
    \end{tabular}
    \caption{Relation types of retrieved commonsense knowledge triples. We use ``Others" to uniformly denote other relations whose proportion is less than 1\%. }
    \label{relation_types}
\end{table}

As shown in Table \ref{relation_types}, among these extracted knowledge triples, there are mainly 6 types of relations, where ``Synonymy" and ``RelatedTo" contribute to the largest proportion, with 41\% and 38\% respectively. 

\section{Model Description}
In this section, we will describe our proposed question generation model, as is illustrated in Figure \ref{model}. Based on the extracted knowledge triples which can provide commonsense transition information, we incorporate this knowledge into question generation via multi-task learning by employing two triple-based auxiliary tasks.

\subsection{Multi-task Learning Framework}
For the commonsense knowledge triple $\mathbf{(h,r,t)}$, where the head concept $\mathbf{h}$ appears in the context passage and tail concept $\mathbf{t}$ appears in the question, it directly provides the conversion information needed for QG. To help the main QG model have a deeper understanding of this commonsense transition, we design two auxiliary tasks: Relation Classification (RC) and Tail Concept Generation (TG). We describe our main QG model and two auxiliary tasks, as well as the unified model in the following parts.

\subsubsection{Main Task: QG Baseline Model}
Given a context passage $p$ and a specific answer $a$, QG targets to generate a grammatical question that can be answered by $a$ based on the content of $p$. We perform sequence-to-sequence generation and adopt the model proposed by Zhang \shortcite{Zhang2019AddressingSD} as our main QG model.

First, we employ a two-layer bi-directional LSTMs as the encoder, which takes feature-enriched embedding $e_i$ as input and outputs a list of hidden representations $H$:
\begin{align}
    H_i &= [\overrightarrow{h_i};\overleftarrow{h_i}]\\
    \overrightarrow{h_i} &= \overrightarrow{LSTM}([e_i;\overrightarrow{h_{i-1}}])\\
    \overleftarrow{h_i} &= \overleftarrow{LSTM}([e_i;\overleftarrow{h_{i-1}}])\\
    e_i &= [w_i;a_i;n_i;p_i]
    \label{encoder}
\end{align}
where $w_i$, $a_i$, $n_i$, $p_i$ respectively represents the embedding of words, answer position (BIO), Name Entity (NER) and Part-of-Speech (POS). For word embedding, we follow the settings of Zhang \shortcite{Zhang2019AddressingSD} and use ELMo \cite{Peters2018DeepCW} or BERT \cite{Devlin2019BERTPO} to obtain contextualized word representations.

To aggregate long-term dependencies within the context passage, we add a gated self-attention mechanism to the encoder outputs $H$ for $\hat{H}$:
\begin{align}
    \hat{h^p_i} &= g_i*f^p_i + (1-g_i)*h^p_i
\end{align}
We obtain self-attention context vector $f^p_i$ through self-matching mechanism on $H$ and then use a learnable gate $g_i$ to balance how much $f^p_i$ and $h^p_i$ will contribute to the output $\hat{H}$.

The decoder we used is a two-layer uni-directional LSTM. At each decoding step $t$, the decoder state $s_t$ is updated dynamically by an attention mechanism on $\hat{H}$:
\begin{align}
    s_{t+1} &= LSTM([y_t,\Tilde{s_t}])\\
    \Tilde{s_t} &= tanh(W^e[c_t;s_t])\\
    c_t &= \hat{H}\alpha_t, \alpha_t = softmax(\hat{H}^TW^hs_t)
\end{align}

For each target word $y_t$, its probability generated from vocabulary is computed by a maxout neural network and softmax function:

\begin{align}
    \hat{u_t} &= tanh(W^d[c_t;s_t])\\
    u_t &= [max\{\Hat{u}_{t,2k-1},\hat{u}_{t,2k}\}]_k\\
    P_{vocab} &= softmax(W^ou_t)
\end{align}

Besides, the pointer mechanism will also be applied to calculate the probability of copying a word from the source context. Finally, the probability distribution is a combination of these two modes with a gate $p_g$:
\begin{align}
    P(y_t|y_{<t}) = p_gP_{vocab} + (1-p_g)P_{copy}
\end{align}

\begin{figure*}[t]
    \centering
    \includegraphics[width=0.98\columnwidth]{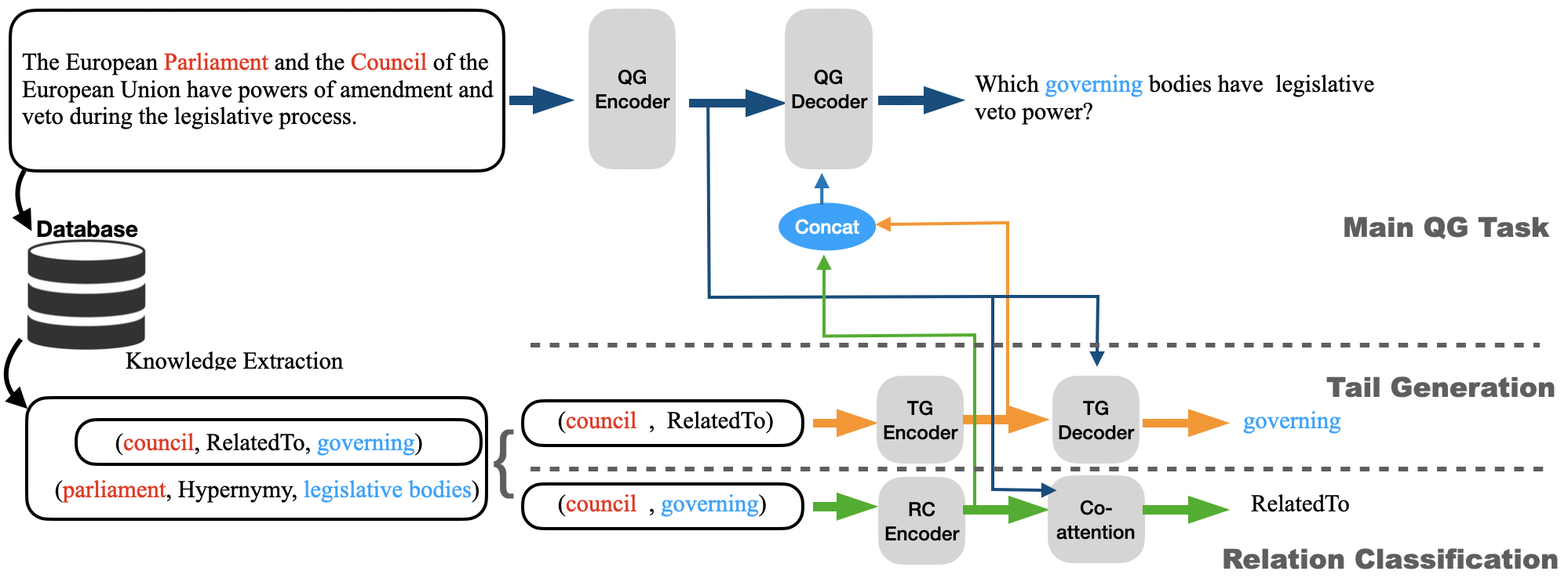}
    \caption{The illustration of our proposed QG framework. First, we conduct knowledge extraction. In the training stage, the two triple-based auxiliary tasks provide commonsense knowledge for the main QG model and the QG model also serves as a context for TG and RC tasks.}
    \label{model}
\end{figure*}

The training objective is to minimize the negative log likelihood of the target sequence $\mathbf{q}$:
\begin{align}
    \mathcal{L}_q = -\frac{1}{T_q}\sum_{t=1}^{T_q}log(P(y_t=\mathbf{q}_t))
\end{align}

\subsubsection{Auxiliary Task-1: Relation Classification}
This task is designed to predict the correct relationship between the head concept $\mathbf{h}$ and tail concept $\mathbf{t}$. We use a two-layer bi-directional LSTM to encode the $\mathbf{(h,t)}$ pair and obtain the hidden representation $R$. Then we conduct co-attention mechanism between $R$ and the context passage representation $\hat{H}$ to get the co-dependent context $\hat{R}$:
\begin{align}
    \hat{R} &= [\hat{H};RA^H]A^R\\
    R &= LSTM([\mathbf{h;t}])\\
    A^H &= softmax((R^T\hat{H})^T)\\
    A^R &=softmax(R^T\hat{H})
\end{align}

Based on $\hat{R}$, we use a feed-forward layer $f$ and softmax function to predict the class of relationship:
\begin{align}
    y_r &= softmax(f(\hat{R}))\\
    \mathcal{L}_r &= -\sum_{r}(\hat{y_r}\log(y_r))
\end{align}
where $\mathcal{L}_r$ is the loss function, and $\hat{y_r}$ is the one-hot label of the relationship class as listed in Table \ref{relation_types}.

\subsubsection{Auxiliary Task-2: Tail Concept Generation}
Correspondingly, given the head concept $\mathbf{h}$ and relationship $\mathbf{r}$, generating a proper tail concept $\mathbf{t}$ also needs a deep understanding of the commonsense knowledge between them, so we design a second auxiliary task: Tail Concept Generation. Specially, we adopt another two-layer bi-directional LSTM to encode $\mathbf{(h,r)}$:
\begin{align}
    T &= LSTM([\mathbf{h;r}])
\end{align}

In the decoding process, we use a uni-directional LSTM to generate tail concept words sequentially based on the head-relation pair. Additionally, QG passage can serve as a background so we add the context representation of QG passage $\hat{H}$ (computed in Equation 5) into tail concept decoding:
\begin{align}
    s_{j+1} &= LSTM([y_j,\Tilde{s_j}])\\
    \Tilde{s_j} &= tanh(W^t[c_j;k_j;s_j])\\
    c_j &= \hat{H}\alpha_j, \alpha_j = softmax(\hat{H}^TW^hs_j)\\
    k_j &= T\gamma_j, \gamma_j = softmax(T^TW^ks_j)
\end{align}
where $y_j$ refers to the $j$-th tail concept word, $c_j$ and $k_j$ represent the context of QG passage and head-relation pair, respectively. The word probability calculation is same with QG model. The loss function of tail concept generation is:
\begin{align}
    \mathcal{L}_t = -\frac{1}{T_t}\sum_{j=1}^{T_t}log(P(y_j=\mathbf{t}_j))
\end{align}

\subsubsection{Unified Model}
Our unified model combines the main QG model and two auxiliary tasks, as illustrated in Figure \ref{model}. In detail, the QG context passage, the head-tail pair and head-relation pair will firstly be encoded by QG encoder, head-tail encoder and head-relation encoder, respectively. Then we concatenate the head-relation and head-tail encoder outputs to obtain the complete knowledge triple representation:
\begin{align}
    K &= concat([T;R])\\
    k_t &= K\beta_t, \beta_t = softmax(K^TW^hs_t)
\end{align}
We use this representation as additional commonsense context for QG decoding process. Therefore, the Equation 7 can be rewritten as:
\begin{align}
    \Tilde{s_t} &= tanh(W^e[c_t;k_t;s_t])
\end{align}
The overall training objective is the combination of the main QG task and two auxiliary tasks:
\begin{align}
    \mathcal{L} =\mathcal{L}_q+\mathcal{L}_r+\mathcal{L}_t
\end{align}

As we mentioned above, each knowledge-equipped training sample has 1.7 corresponding triples on average. Since our proposed auxiliary tasks only take one knowledge triple as input at a time, we need to choose one triple among several extracted triples. As shown in Table \ref{relation_types}, the types of “synonymy” and “relatedto” have the largest proportions in all extracted triples. We test the effects of prioritizing “synonymy” triples and prioritizing “relatedto” triples. According to experiment results, priority use of “synonymy” is slightly better than the priority use of “relatedto” triples. Based on this, we prefer to use “synonymy” triples in our models.

\subsection{Iterative Training Framework}
Actually, only 61.3\% percent of the training samples have extracted knowledge triples, and our unified model can only be trained on these knowledge-equipped samples. In order to make full use of the remaining pure training samples, we adopt an iterative training framework (ITF) that alternately utilizes knowledge-triple-equipped training data and pure training data.

Our unified model is composed of three parts: QG model, RC model and TG model, where RC and TG models serve as auxiliary tasks to provide commonsense knowledge for QG. During iterative training, the unified model is firstly trained based on knowledge-triple-equipped training data for $N$ steps. Then we switch to the pure training data for another $N$ steps and only update the QG model's parameters and leave the other parameters related to using knowledge triples frozen. In this way, the pure training samples can also contribute to the unified model.

\section{Experimental Settings}

\subsection{Dataset and Metrics} As the most commonly used QG dataset, SQuAD is composed of (passage, answer, question) samples. Follow the setting of Zhang \shortcite{Zhang2019AddressingSD}, we split the accessible parts of SQuAD into a training set (75,722 samples), a development set (10,570 samples) and a test set (11,877 samples).

We evaluate the performance of our models using BLEU \cite{papineni-etal-2002-bleu}, ROUGE-L \cite{Lin2004ROUGEAP} and METEOR \cite{Denkowski2014MeteorUL}.

\subsection{Baseline Models}
We compare our method with the following previous works on SQuAD.
\begin{itemize}
    \item \textbf{Rule-based methods:} PCFG-Trans \cite{Heilman2010GoodQS}, Syn-QG \cite{Dhole2020SynQGSA}
    \item \textbf{Pre-trained models:} ACS-QG \cite{Liu2020AskingQT}, UNILM \cite{Wang2020MiniLMDS}, ERNIE-GEN \cite{Xiao2020ERNIEGENAE}, UNILMv2 \cite{Bao2020UniLMv2PL}, ProphetNet \cite{Yan2020ProphetNetPF}
    \item \textbf{Seq2Seq models:} NQG++ \cite{Zhou2017NeuralQG}, M2S+cp \cite{Song2018LeveragingCI}, A-P-Hybrid \cite{Sun2018AnswerfocusedAP}, s2s-a-ct-mp-gsa \cite{Zhao2018ParagraphlevelNQ}, Q-type \cite{zhou-etal-2019-question}, sent-Relation \cite{Li2019ImprovingQG}, Capture Great Context \cite{Luu2020CapturingGC}, NQG-RL-GS \cite{Chen2019NaturalQG}, QPP\&QAP \cite{Zhang2019AddressingSD}, Paraphrase-QG \cite{Jia2020HowTA}
    
\end{itemize}

Besides, in order to present the different performance between our multi-task learning framework of using knowledge triples and previous methods, we also implement the graph attention method \cite{Zhou2018CommonsenseKA}, which is proposed to facilitate conversation generation task, as a comparing model. This method constructs the retrieved triples into a graph and attentively reads the knowledge triples within each graph through a dynamic graph attention mechanism.
\begin{table*}[ht]
    \centering
    \scalebox{0.8}{
    \begin{tabular}{c|c|cccccc}
    \hline
         Categories&Models & BLEU-1 & BLEU-2 & BLEU-3 & BLEU-4 & ROUGH-L & METEOR \\
    \hline
    \hline
    \multirow{2}{*}{Rule-based}&PCFG-Trans& 28.77&17.81&12.64&9.47&31.68&18.97\\
    &Syn-QG & \bf45.55&\bf30.24&\bf23.84&\bf18.72&-&-\\
    \hline
    \multirow{5}{*}{Pre-trained}&ACS-QG&\bf52.30& \bf36.70& \bf28.00& 22.05&53.25&25.11\\
    &UNILM& -&-&-&23.75&52.04&25.61\\
    &ERNIE-GEN&-&-&-&25.57&53.31&26.89\\
    &UNILMv2&-&-&-&26.30&53.19&27.09\\
    &ProphetNet&-&-&-&\bf26.72&\bf53.79&\bf27.64\\
    \hline
    \multirow{15}{*}{Seq2Seq}&NQG++&42.36&26.33&18.46&13.51&41.60&18.18\\
    &M2S+cp&-&-&-&13.91&42.72&18.77\\
    &A-P-Hybrid&43.02&28.14&20.51&15.64&-&-\\
    &s2s-a-ct-mp-gsa&44.51&29.07&21.06&15.82&44.24&19.67\\
    &Q-type&43.11&29.13&21.39&16.31&-&-\\
    &Sent-Relation&44.40&29.48&21.54&16.37&44.73&20.68\\
    &Paraphrase-QG&44.32&29.88&22.28&17.21&-&20.96\\
    &Capture Great Context&\bf46.60&\bf31.94&23.44&17.76&45.89&21.56\\
    &NQG-RL-GS&-&-&-&17.94 &46.02&21.76\\
    &QPP\&QAP&-&-&-&18.65&\bf46.76&\bf22.91\\
    \cline{2-8}
    &QG baseline model (with ELMo)&44.99&30.03&22.05&16.70&45.15&21.11\\
    &+ graph attention &45.34&30.18&22.29&17.14&45.04&20.72\\
    & Our Unified model (EMLo) &45.44&30.64&22.63&17.31&46.02&21.58\\
    \cline{2-8}
    &QG baseline model (with BERT)&46.12&31.41&23.51&18.19&46.41&21.69\\
    & Our Unified model (BERT) &46.36&31.74&\bf23.91&\bf 18.65&46.65&21.84\\
    \hline
    \end{tabular}
    }
    \caption{Experimental results of our unified model comparing with previous works}
    \label{whole_experiment}
\end{table*}

\subsection{Implementation Details}
Following the settings of Zhang \shortcite{Zhang2019AddressingSD}, we tokenize and obtain POS/NER features by Stanford Corenlp\footnote{http://stanfordnlp.github.io/CoreNLP/}. The QG encoder, TG encoder, RC encoder, QG decoder, and TG decoder are all 2-layer LSTMs with the hidden size of 600. We set the probability of dropout to 0.3 for each layer. We use beam search with size of 10 for decoding. In the Iterative Training Framework (ITF), we set $N$ to 3000 and each training mode will be iteratively trained 3 times. In order to reduce the volatility of the training process, we averaged the 5 models closest to the best performing checkpoint on the development set.

The pre-trained ELMo \cite{Peters2018DeepCW} word embedding is character-level and we keep it fixed during training. For the BERT \cite{Devlin2019BERTPO} version model, we utilize BERT embeddings as the replacement of ELMo. WordPiece tokenizer is applied to tokenize each word and POS/NER tags are also extended to its corresponding word pieces. During inference, we map the word-piece outputs to normal words through post-processing. 
\section{Results}
\subsection{Main Results}
The main experimental results are shown in Table \ref{whole_experiment}. For fair comparisons, we divide previous works into three categories: rule-based methods, pre-trained language modeling-based methods and Seq2Seq methods, where rule-based methods and pre-trained models only serve as references and different Seq2Seq models are our comparing methods.    

Our unified model (with BERT) outperforms all but one previous Seq2Seq models and obtains comparable results (18.65 BLEU-4 score) with the best QPP\&QAP method \cite{Zhang2019AddressingSD}. QPP\&QAP method relies on two pre-trained models: question paraphrasing classification and question answering models to provide rewards for policy gradient, and is further equipped with reinforcement learning mechanism, which is more complicated than ours.

Compared with the best performance of rule-based method Syn-QG, our evaluation score of BLEU-4 is very close while our other scores are obviously better.

Although the performance of our model is still far from the methods that are based on pre-training language modeling, our method is much simpler and provides an effective way to use diverse knowledge databases to supplement knowledge triples into QG. Our idea of introducing knowledge into the generation process is completely different from pre-trained models. Actually, pre-training turns out to be a useful strategy for helping models with a large number of parameters based on large-scale unsupervised training corpus \cite{Dong2019UnifiedLM,Liu2020AskingQT,Xiao2020ERNIEGENAE,Bao2020UniLMv2PL,Yan2020ProphetNetPF}. However, it is extremely time-consuming and computationally expensive. 

Compared with these, our method is lightweight and only takes several related knowledge triples as additional inputs instead of huge amounts of data. 

Besides, directly using top $K$ extracted triples\footnote{We set $K$ to 3*$m$ in our experiments, where $m$ represents the number of content words in each paragraph.} as additional inputs through graph attention \cite{Zhou2018CommonsenseKA} has a slight improvement over the QG baseline model (17.14 vs. 16.70 on BLEU-4). Compared with this traditional method, our proposed multi-task learning framework has better performance, achieving a 0.61 increase over the QG baseline model (with ELMo). Meanwhile, after applying our proposed framework, we also achieve a 0.46 BLEU-4 improvement over the much stronger QG baseline model (with BERT), which demonstrates that our proposed methods can indeed improve the performance of question generation.

\begin{table*}[ht]
    \centering
    \scalebox{0.9}{
    \begin{tabular}{l|cccccc}
    \hline
         Model & BLEU-1 & BLEU-2 & BLEU-3 & BLEU-4 & ROUGH-L & METEOR \\
    \hline
    \hline
        QG baseline model & 44.47&29.28 &21.23 &15.90 &44.30 &20.52 \\
    \hline
        Unified model (w.o. ITF) &44.91&29.96&\bf21.96&\bf16.67&\bf45.48&	\bf21.28\\
        \quad -TG & 44.30&	29.47&	21.60&	16.32&	45.10&	20.70\\
        \quad -RC&\bf46.21&	\bf30.37&	21.95&	16.38&	44.04&	21.10\\
        \quad -TG-RC & 45.04&29.71 &21.57 &16.15 &44.84 &20.77 \\
    \hline
    \end{tabular}
    }
    \caption{Ablation studies of our unified model (with ELMo) trained on \textbf{knowledge-equipped data}.}
    \label{detailed_experiment}
\end{table*}
\subsection{Ablation Study}
We perform model variants and conduct ablation tests for better understanding the effect of different components of our model, 
and the results are shown in Table \ref{detailed_experiment}.

Through our Iterative Training Framework (ITF), we can simultaneously utilize the knowledge-equipped training data and pure training data.
In order to better displaying the improvement brought by the introduction of external commonsense knowledge alone, we also conduct experiments by training the model only on the \textbf{knowledge-equipped training data}. 
In this case, as shown in Table \ref{detailed_experiment},
the QG baseline model (with ELMo) only obtains a 15.90 BLEU-4 score. After applying our proposed components, our unified model boosts the BLEU-4 score to 16.67. That means, with the help of extracted commonsense knowledge and our multi-task learning framework, 61.3\%   training data (as listed in Table \ref{triple_statistics}) yields very close results to the performance of training on the whole SQuAD dataset (16.70 in Table \ref{whole_experiment}).

\begin{table*}[t]
    \centering
    \scalebox{0.8}{
    \begin{tabular}{p{0.9\columnwidth}}
    \hline
    \hline
    \textbf{Context Passage-1:} \\
    Gateway National Recreation Area contains over 26,000 acres
    ( \underline{\textbf{10,521.83}} {\color{red}{ha}} ) in total , most of it surrounded by New York City , including the Jamaica Bay Wildlife Refuge in Brooklyn and Queens , over 9,000 acres ( 36 km2 ) of salt marsh , islands , and water , including most of Jamaica Bay . Also ......\\
    \textbf{Answer:}\\
    10,521.83 \\
    \textbf{Reference:} \\
    how large is the gateway national recreation area in {\color{red}{hectares}} ?\\
    \textbf{Baseline:}\\
    how many acres contains gateway national recreation area ?\\
    \textbf{Unified:}\\
    how many {\color{red}{hectares}} does the gateway national recreation area have?\\
    \hline
    \hline
    \textbf{Context Passage-2:}\\
    Australia : The event was held in Canberra , Australian Capital Territory on April 24 , and covered around 16 km of Canberra 's central areas , from Reconciliation Place to Commonwealth Park . Upon its arrival in Canberra , the Olympic {\color{red}{flame}} was presented by Chinese officials to local Aboriginal elder \underline{\textbf{Agnes Shea}} , of the Ngunnawal people . She , in turn ......\\
    \textbf{Answer:}\\
    Agnes Shea\\
    \textbf{Reference:}\\
    what is the name of the aboriginal elder who received the {\color{red}{torch}} from chinese officials ? \\
    \textbf{Baseline:}\\
    who was the olympic flame presented to ?\\
    \textbf{Unified:}\\
    who received the {\color{red}{torch}} from chinese officials ?\\
    
    \hline
    \hline
    \end{tabular}}
    \caption{Two real cases in the test set of SQuAD. We bolded the answer and highlight the parts of the passage and the question that require commonsense knowledge conversion. The baseline model refers to QG baseline model (with BERT) and Unified model refers to Unified model (BERT) with ITF. }
    \label{case_study}
\end{table*}
To confirm the effect of each component we proposed, we conduct ablation experiments over the unified model based on the \textbf{knowledge-equipped training data}. Without the Tail Generation auxiliary task, the performance of our unified model has a drop of 0.35. Besides, the unified model has a performance degradation of 0.29 if removing the Relation Classification task. In the case of removing two auxiliary tasks at the same time (using knowledge triple as an  additional input for attention mechanism), the effect of the model will drop by 0.52. These experimental results verify the effectiveness of each auxiliary task and the combination of them.

\subsection{Analysis of Auxiliary Tasks}
In addition to the main QG task, we also evaluate the performance of two triple-based auxiliary tasks. The Relation Classification task has six categories and its accuracy reaches 66\%, which has a 25\% increase over the most-frequent-category baseline (41\%). For RC task, the relationship between the head and tail concept is generally unique.

On the contrary, Tail Concept Generation is a much difficult task because given head concept and relation, the tail concept is not unique. We use BLEU-1 as evaluate metric and obtain a 6.23 score on this task.

\subsection{Human Evaluation}
For a text generation task, the automatic metrics like BLEU, ROUGH, and METEOR have limitations to evaluate the quality of generated questions. Therefore, we conduct human evaluation to compare the performance of Unified model (BERT) and QG baseline model (with BERT). We randomly select 100 samples and ask three annotators to score the generated questions of two models, according to: \textit{\bf Relevancy}: whether the question is relevant to the context passage; \textit{\bf Fluency}: whether the question is grammatical and fluent; \textit{\bf Answerability}: whether the question can be answered by the given answer.
The rating score is set to [0, 2]. The evaluation results are shown in Table \ref{human-eval}. Our unified model receives higher scores on all three metrics. Moreover, the high Spearman correlation coefficients guarantee the validity of our human evaluation results. 

\begin{table}[ht]
    \centering
    \scalebox{0.85}{
    \begin{tabular}{c|ccc}
    \hline
         Model&Relevancy&Fluency&Answerability  \\
    \hline
    \hline
         baseline/unified& 1.39/1.41&1.74/1.80&1.44/1.47\\
    \hline
         Spearman&0.65&0.80&0.73\\
    \hline
    \end{tabular}
    }
    \caption{Human evaluation results.}
    \label{human-eval}
\end{table}
\subsection{Case Study}
To clearly display the output questions, two real cases in the test set of SQuAD are shown in Table \ref{case_study}. Generating the right questions needs commonsense knowledge that cannot be directly obtained from the given passage, such as ``ha" is the short form of ``hectares" in case-1 and ``flame" is synonymy with ``torch" in case-2. For the baseline model, lacking such commonsense knowledge results in unsatisfactory questions which only copy some words from the passage. Compared with the baseline model, our unified model can better deal with these cases with the help of external commonsense knowledge.

\section{Conclusion}
In this paper, we propose a new multi-task learning framework to introduce commonsense knowledge into QG. We first extract relevant structured knowledge triples from external databases, ConceptNet and WordNet. Based on these knowledge triples, we design two auxiliary tasks to help the main QG model deeply absorb the commonsense knowledge. Both the automatic and human evaluation results verify the effectiveness of our proposed methods. In the future, we will explore new ways to use multiple knowledge triples simultaneously in the multi-task learning framework. Besides, we may also apply our framework in other text generation tasks, such as conversation generation and story generation.

\section*{Acknowledgments}
This  work  is  supported  by  the  National  Natural  Science  Foundation  of  China  (62076008, 61773026)  and  the  KeyProject of Natural Science Foundation of China (61936012).

\bibliographystyle{ccl}
\bibliography{submission}

\end{document}